\pdfoutput=1
\documentclass[11pt]{article}

\usepackage[final]{acl}

\usepackage{times}
\usepackage{latexsym}

\usepackage[T1]{fontenc}
\usepackage[utf8]{inputenc}

\usepackage{microtype}

\usepackage{inconsolata}
    
\usepackage{amsmath}

\usepackage{amsfonts}

\usepackage{svg}
\svgpath{{figures/}} %set the default svg path

\usepackage{bbm}

\usepackage{amsthm}

\usepackage{adjustbox}

\usepackage{subcaption}
\usepackage{caption}

\usepackage{multirow}

\usepackage{soul}

\usepackage{lipsum}

\title{Unsupervised Parsing by Searching for Frequent Word Sequences among Sentences with Equivalent Predicate-Argument Structures}

\author{Junjie Chen \\
  The University of Tokyo \\
  \texttt{christopher@is.s.u-tokyo.ac.jp} \\\And
  Xiangheng He \\
  Imperial College London \\
  \texttt{x.he20@imperial.ac.uk} \\\AND
  Danushka Bollegala \\
  The University of Liverpool \\
  \texttt{danushka@liverpool.ac.uk        } \\\And
  Yusuke Miyao \\
  The University of Tokyo\\
  \texttt{yusuke@is.s.u-tokyo.ac.jp}}

\begin{document}
\maketitle
% \vspace{10cm}
% \input{acl_instruction.tex}
\begin{abstract}
    Unsupervised constituency parsing focuses on identifying word sequences that form a syntactic unit (i.e., constituents) in target sentences.
    Linguists identify the constituent by evaluating a set of Predicate-Argument Structure (PAS) equivalent sentences where we find the constituent appears more frequently than non-constituents (i.e., the constituent corresponds to a frequent word sequence within the sentence set).
    However, such frequency information is unavailable in previous parsing methods that identify the constituent by observing sentences with diverse PAS.
    In this study, we empirically show that \textbf{constituents correspond to frequent word sequences in the PAS-equivalent sentence set}.
    We propose a frequency-based parser \emph{span-overlap} that (1) computes the span-overlap score as the word sequence's frequency in the PAS-equivalent sentence set and (2) identifies the constituent structure by finding a constituent tree with the maximum span-overlap score.
    The parser achieves state-of-the-art level parsing accuracy, outperforming existing unsupervised parsers in eight out of ten languages.
    Additionally, we discover a multilingual phenomenon: participant-denoting constituents tend to have higher span-overlap scores than equal-length event-denoting constituents, meaning that the former tend to appear more frequently in the PAS-equivalent sentence set than the latter.
    The phenomenon indicates a statistical difference between the two constituent types, laying the foundation for future labeled unsupervised parsing research.
\end{abstract}

\section{Introduction}

\begin{figure}[t]
    \centering
    \includesvg[inkscapelatex=false, width=0.8\columnwidth]{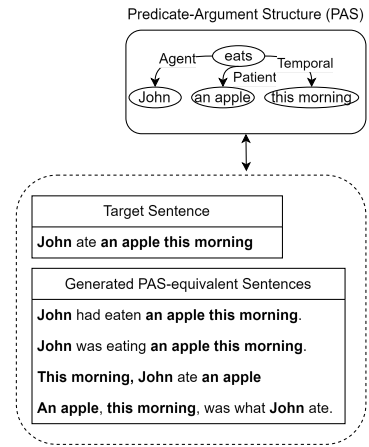}
    \caption{A target sentence and a set of PAS-equivalent sentences. We highlight in bold constituents that are frequent in the PAS-equivalent sentence set.}
    \label{fig:thematic_equiv_example}
    % \vspace{-0.6cm}
\end{figure}

\newcommand\blfootnote[1]{
  \begingroup
  \renewcommand\thefootnote{}\footnote{#1}%
  \addtocounter{footnote}{-1}%
  \endgroup
}

\blfootnote{We release the code and data at \url{https://github.com/junjiechen-chris/ACL24-SpanOverlap}}
Unsupervised constituency parsing focuses on identifying the constituent structure for target sentences.
The constituent structure consists of a set of syntactically self-contained word sequences, commonly known as constituents.
In addition to being syntactic units, those constituents encode semantic units in the sentence's semantic structure \cite{Carnie2007-CARSAG, Fasold_Connor-Linton_2014}.
Among all possible word sequences that can arise in natural language, the constituent is one of the few word sequences that can describe the semantic unit.
As a result, we would have a higher chance of repeatedly observing the constituent than random word sequences given the semantic unit.
In other words, we can observe the constituent as a frequent word sequence among semantically equivalent sentences.
We refer to the frequent word sequence as a \emph{word sequence pattern}.

In this paper, we empirically show a hypothesis: \textbf{constituents correspond to frequent word sequences in the set of sentences with equivalent Predicate-Argument Structures (PAS)}.
PAS abstracts the sentence semantics as an event with three elements:  (1) predicates of the event; (2) participants of the event (i.e., arguments); (3) the semantic relationship between the predicates and arguments (i.e., thematic relations) \cite{palmer-etal-2005-proposition} (Figure~\ref{fig:thematic_equiv_example}).
We refer to two sentences as \emph{Predicate-Argument Structure (PAS) equivalent} when they have the same PAS elements.
Controlling the PAS provides a good approximation to controlling the semantic structure.
Figure~\ref{fig:thematic_equiv_example} exemplifies our hypothesis with a target sentence, ``John ate an apple this morning.''
Word sequences like ``an apple'' are constituents in the target sentence, whereas word sequences like ``John ate'' are non-constituents.
Compared to the non-constituent, the constituent appears more frequently in the target sentence's PAS-equivalent sentence set, manifesting itself as a word sequence pattern.

\begin{table}[t]
    \adjustbox{max width=1\columnwidth}{
        \centering
        % \small
        \begin{tabular}{|l|l|}
            \hline
            \multirow{4}{*}{Penn Treebank           }   & Results were released after the market closed.      \\
             & Markets stopped trading in many securities. \\
             & Mr. Bandow is a Cato Institute fellow.      \\\hline
            \multirow{4}{*}{Constituency Test  } & It's an apple that John eats this morning.          \\
             & What John eats this morning is an apple.    \\
             & An apple, John eats this morning            \\\hline
        \end{tabular}
    }
    \caption{A comparison of data in Penn Treebank and data produced by constituency tests. The constituency test data are byproducts of testing the constituency of ``an apple''.}
    \label{tbl:treebank_vs_linguistics}
    % \vspace{-0.4cm}

\end{table}

However, such frequency information was not available to previous unsupervised parsing methods.
Those previous methods focused on uncovering the constituent structure by either inducing a complex grammar \cite{kim-etal-2019-compound, yang-etal-2021-pcfgs} or evaluating whether a word sequence can withstand syntactic transformations \cite{cao-etal-2020-unsupervised, wu-etal-2020-perturbed, li-lu-2023-contextual}.
Both methods operate upon sentences with diverse PAS, as shown in the upper half of Table~\ref{tbl:treebank_vs_linguistics}.
These sentences provide little information about the constituent structure, as evidenced by the low frequency of word sequences corresponding to constituents.
Linguistic researchers perform constituency tests (e.g., movement) \cite{Carnie2007-CARSAG} using handcrafted sentences with equivalent PAS, as shown in the lower half of Table~\ref{tbl:treebank_vs_linguistics}.
These PAS-equivalent sentences, as discussed above, provide significantly more information about the constituent structure, as evidenced by the high frequency of word sequences corresponding to constituents.
The gap in data characteristics motivates our research on unsupervised parsing using PAS-equivalent sentences.

We apply the hypothesis to unsupervised parsing and propose a frequency-based method, \emph{span-overlap}, to exploit the frequency information.
The span-overlap method (1) generates a PAS-equivalent sentence set using a Large Language Model (LLM), (2) computes the span-overlap score for word sequences as the sequence's frequency in the sentence set, and (3) identifies the constituent structure by finding the constituent tree with maximum span-overlap scores.
The span-overlap parser obtains higher parsing accuracy than state-of-the-art unsupervised parsers in eight out of ten languages evaluated and by a large margin.
The span-overlap score can effectively separate constituents from non-constituents, especially for long constituents.
These results highlight the utility and applicability of the frequency information to unsupervised parsing.
In addition to the hypothesis, we discover a multilingual phenomenon: participant-denoting constituents tend to have higher span-overlap scores than equal-length event-denoting constituents, meaning that the former tend to appear more frequently in the PAS-equivalent sentence set than the latter.
The phenomenon indicates a statistical difference between the two constituent types, laying the foundation for labeled unsupervised parsing research.

\section{Background}
Unsupervised constituency parsers seek to identify unlabeled constituents for any given sentence.
A constituent in a sentence $w:=(w[1], ..., w[n])$ is a consecutive word sequence $w^{i, j}:=(w[i], ..., w[j])$ that functions as an independent unit in the structure.
The word sequence can be represented as a two-tuple $(i, j)$, referred to as a span.
We measure the parsing accuracy by the F1 score between the set of predicted spans and the set of gold spans \cite{klein-manning-2002-generative}.

Unsupervised parsers make predictions by (1) evaluating a span score for each word sequence and (2) finding a tree optimizing the span score.
Grammar-based unsupervised parsers \cite{yang-etal-2021-pcfgs, zhao-titov-2020-visually} compute the span score as the posterior span probability $p_\mathcal{G}((i, j)|w)$ with respect to a probabilistic grammar $\mathcal{G}$ (i.e., how likely the grammar selects $(i, j)$ as a constituent).
The grammar-based parsers predict constituents by maximizing the span score, seeking to maximize the probability of the prediction containing constituents.
Language model-based parsers \cite{cao-etal-2020-unsupervised, wu-etal-2020-perturbed, li-lu-2023-contextual} compute the span score by whether the language model reacts to syntactic transformations involving the span.
The language model-based parsers predict constituents by minimizing the span score, searching for constituents most independent of the remainder of the sentence.

The language model-based unsupervised parsing methods have attracted much attention lately due to the language model's applicability in various syntactic \cite{hu-etal-2020-systematic, futrell-etal-2019-neural} and semantic tasks \cite{huang-chang-2023-towards}.
Given a word sequence $w^{i, j}$, the method evaluates the span score as the distance $d$ between $w$ (the original sentence) and $f(w, (i, j))$ (the transformed sentence designed to test the constituency of $w^{i, j}$), as shown in Equation~\ref{eq:lm-scoring}.
\citet{cao-etal-2020-unsupervised} measures the distance by the change in the acceptability between $w$ and $f(w, (i, j))$ using an external acceptability model trained with synthetic data.
\citet{wu-etal-2020-perturbed} and \citet{li-lu-2023-contextual} measure the difference by the distance in language model embeddings between $w$ and $f(w, (i, j))$.
\begin{equation}
    {\rm score}(w, (i, j)):=d(w, f(w, (i, j)))
    \label{eq:lm-scoring}
\end{equation}

Recent advances in Large Language Models (LLM) have led to a surge in the application of LLM-enhanced data to Natural Language Processing systems \cite{yang-etal-2023-distribution, perez-etal-2023-discovering, liu-etal-2022-wanli}.
Given an instruction (typically in the form of prompts), the LLM iteratively generates its answer based on the instruction and partially generated answers.
Recent studies \cite{DBLP:journals/corr/abs-2311-07911, DBLP:journals/corr/abs-2306-04757} point out that state-of-the-art LLMs can follow difficult instructions and generate the correct answer with over 90\% accuracy.
This excellent instruction-following capability enables us to sample PAS-equivalent sentences from the LLM.

\section{Related Works}
\label{sec:related-works}
Grammar-based unsupervised parsers \cite{klein-manning-2002-generative, kim-etal-2019-compound, zhao-titov-2020-visually} learn probabilistic grammar by maximizing the likelihood of sentences observed in a corpus.
However, these sentences have diverse PAS to the degree that constituent word sequences can only be observed a limited number of times.
The limited observation provides insufficient positive evidence on whether a word sequence is a constituent through the sequence's frequency.
Also, the grammar has to reject non-constituent but frequent word sequences tied to certain expressions.
For example, the word sequence ``it is'' is a non-constituent but frequent word sequence in sentences like ``(it (is (a book)))''.
The lack of positive evidence and the abundance of false-positive evidence make up the main challenges in unsupervised constituency parsing.
The two issues likely contribute to the low performance of the grammar-based method in non-English languages (see Table~\ref{tbl:spmrl-comparison} and Table~\ref{tbl:ptbnctb-comprison}).
Our span-overlap method is immune to the two issues.
Our method detects the constituent-related word sequence pattern, utilizing the positive evidence revealed in the PAS-equivalent sentences.
The PAS-equivalent sentences express the same PAS in diverse ways, effectively suppressing the frequency of the expression-tied non-constituent word sequences (i.e., false-positive evidence).
Our method achieves significantly higher parsing performance than the grammar-based methods in non-English languages.

Previous language model-based methods conduct constituency tests by applying handcrafted templates to word sequences.
However, different languages implement the syntactic transformation differently, requiring templates tailored to each language.
To circumvent the problem, \citet{cao-etal-2020-unsupervised} limit the scope of languages to English, which results in an English-only parser.
\citet{wu-etal-2020-perturbed} and \citet{li-lu-2023-contextual} limit the transformation to movement and standalone tests only, which limits their parser's performance.
By leveraging LLMs for performing the transformation, our span-overlap method can cover a diverse set of transformations while being applicable to many languages.
Furthermore, the previous method has to perform the constituency test for every word sequence, incurring $O(n^2)$ language model evaluations for sentences with $n$ words.
Our method only generates a few dozen acceptable PAS-equivalent sentences for each target sentence, incurring only $O(1)$ evaluations of the LLM.
The low complexity allows for more efficient language model-based parsing.
For example, our method takes 14 minutes to parse the development set of Penn Treebank \cite{Treebank-3} whereas \citet{li-lu-2023-contextual}'s method takes about an hour.

Researchers have long pointed out that language models capture shallow syntax, such as subject-verb agreements \cite{hu-etal-2020-systematic} and phrase segmentations \cite{luo-etal-2019-improving, downey-etal-2022-masked}.
However, the shallow syntax covers only a small subset of information that hierarchical constituent structures provide.
For example, knowing ``a beautiful New York City'' as an NP does not provide any information about the NP's internal structure (i.e., ``(a (beautiful ((New York) City)))'').
Language model-based methods utilize the language model's ability to identify shallow syntax for reconstructing the hierarchical constituent structure.
Our span-overlap method represents an improvement over previous language model-based methods, as discussed in the above paragraph.
As we will see in Section~\ref{sec:experiment}, both our span-overlap parser and the baseline language model-based parser outperform the GPT-3.5 parser in all languages. 
This result indicates that LLM alone cannot recover the full constituent structure, prompting further research on language model-based unsupervised parsing.

\section{Method}
\begin{figure}[t]
    \centering
    \includesvg[inkscapelatex=false, width=0.8\columnwidth]{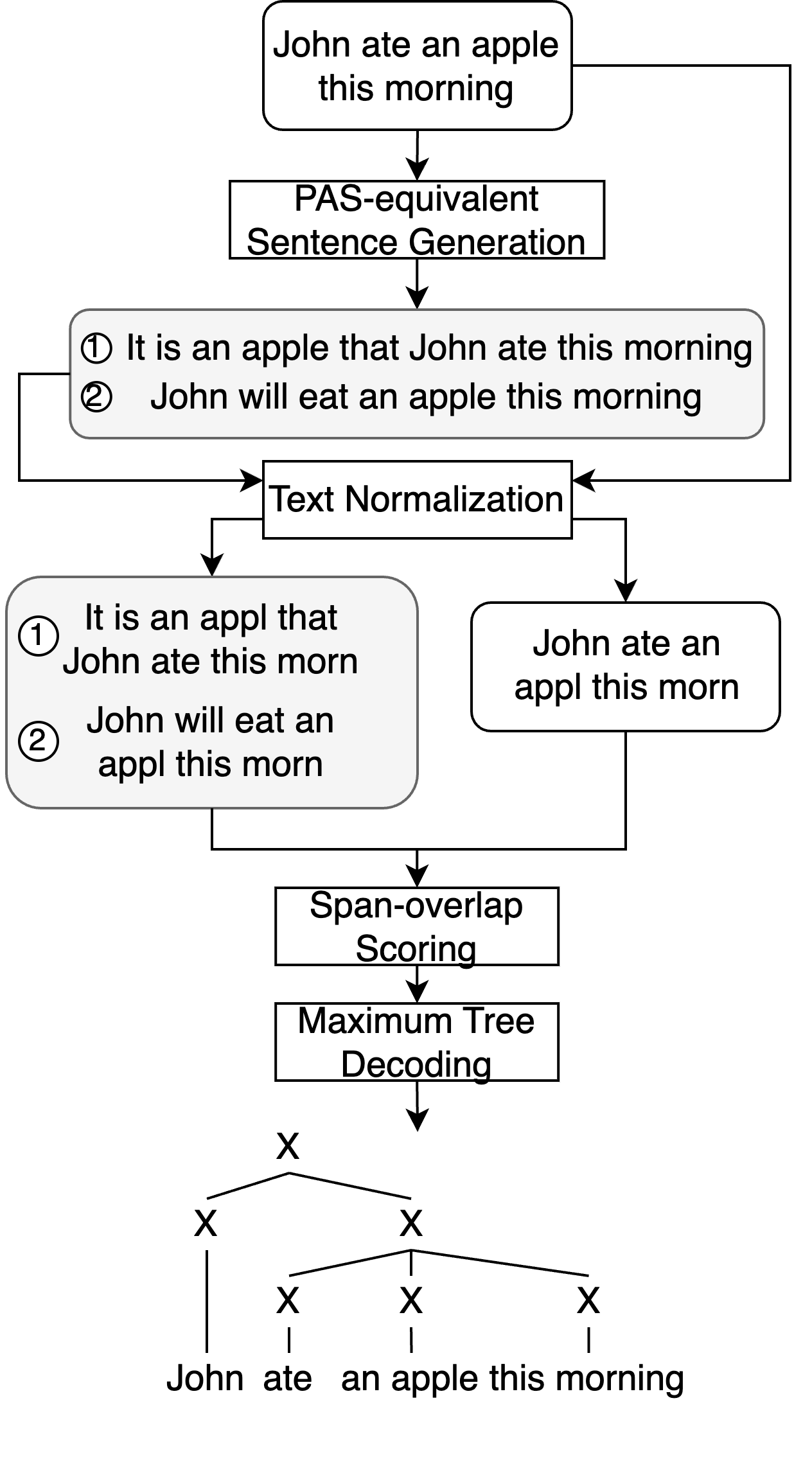}
    % \vspace{-0.6cm}
    \caption{Overview of the span-overlap method. The grey box indicates the set of PAS-equivalent sentences.}
    \label{fig:method-overview}
    % \vspace{-0.4cm}
\end{figure}
% \begin{table*}[t]
%     \centering
%     \tiny
%     \begin{tabular}{|l|l|l|}
%         \hline
%         Original                                        & Instructions                         & Samples                                      \\ \hline
%         \multirow{6}{*}{John ate an apple this evening} & \multirow{1}{*}{Movement}            & This evening, John ate an apple.             \\ \cline{2-3}
%                                                         & \multirow{1}{*}{Question Conversion} & Did John eat an apple this evening           \\ \cline{2-3}
%                                                         & \multirow{1}{*}{Tense Conversion}    & John will eat an apple this evening          \\ \cline{2-3}
%                                                         & \multirow{1}{*}{Clefting}            & It is John who ate an apple this evening     \\ \cline{2-3}
%                                                         & \multirow{1}{*}{Passivization}       & An apple has been eaten by John this evening \\ \cline{2-3}
%                                                         & \multirow{1}{*}{Heavy NP Shift}      & John ate this evening an apple.              \\ \cline{1-3}
%     \end{tabular}
%     \caption{Example of PAS-equivalent sentences produced by the four instructions.}
%     \label{tbl:prompt examples}
%     \vspace{-0.4cm}
% \end{table*}

\begin{table}[t]
    \centering
    \small
    \adjustbox{width=\columnwidth}{
        \begin{tabular}{|l|l|}
            \hline

            \multicolumn{2}{|l|}{Original Sentence }                                            \\\hline
            \multicolumn{2}{|l|}{John ate an apple this evening }                               \\\hline\hline
            % ~                                    & John ate an apple this evening               \\\hline\hline
            Instructions                         & PAS-equivalent Sentences                     \\ \hline
            % \multicolumn{2}{|l|}{PAS-equivalent Sentences}                                      \\\hline
            \multirow{1}{*}{Movement}            & This evening, John ate an apple.             \\ \hline
            \multirow{1}{*}{Question Conversion} & Did John eat an apple this evening           \\ \hline
            \multirow{1}{*}{Tense Conversion}    & John will eat an apple this evening          \\ \hline
            \multirow{1}{*}{Clefting}            & It is John who ate an apple this evening     \\ \hline
            \multirow{1}{*}{Passivization}       & An apple has been eaten by John this evening \\ \hline
            \multirow{1}{*}{Heavy NP Shift}      & John ate this evening an apple.              \\ \hline
        \end{tabular}}
    \caption{Example of PAS-equivalent sentences produced by the six instruction types.}
    \label{tbl:prompt examples}
    % \vspace{-0.4cm}
\end{table}

This section introduces the proposed span-overlap method  (Figure~\ref{fig:method-overview}).
Given a target sentence $w$, the span-overlap method performs parsing in four steps:
(1) Generating a set of sentences $s=\{s_1, s_2, ...\}$ where all $s_k$ are PAS-equivalent to $w$;
(2) Normalizing words in $w$ and $s_k\in s$;
(3) Measuring the frequency of all word sequences of $w$ within the sentence set $s$. We denote each word sequence that extends from the i-th word to the j-th word as $w^{i, j}$. We refer to the frequency of $w^{i, j}$ within $s$ as the \emph{span-overlap score} for the span $(i, j)$;
(4) Performing maximum tree decoding over the span-overlap score.

Firstly, we generate a PAS-equivalent sentence set $s$ for $w$ using an instruction-following LLM (e.g., \texttt{GPT-3.5}).
We instruct the LLM with six types of PAS-preserving instructions listed in Table~\ref{tbl:prompt examples}.
The movement instruction contains four sub-instructions common in most languages (topicalization, preposition, postposition, and extraposition).
The instruction models that constituents can be moved around as an independent unit.
The question conversion instruction asks the LLM to convert $w$ to a yes-no question or a wh-question.
The tense conversion instruction asks the LLM to convert $w$ to another tense.
The instruction models the closer relationship between the verb and object via adding or removing auxiliaries between the verb and subject.
The clefting instruction asks the LLM to construct a cleft sentence based on $w$, modeling subject and object extractions in embedded clauses.
The passivization asks the LLM to convert $w$ from an active voice to a passive voice.
The language-specific instruction contains three sub-instructions: scrambling, ba-construction, and heavy NP shift.
We present the instruction configuration in Appendix~\ref{sec:appendix-instru}.

Secondly, we normalize words in $w$ and $s_k\in s$ to reduce incorrect word sequence mismatches due to unmatching inflectional suffixes or misaligned tokenization settings.
We perform stemming and contraction normalization in this step.
The stemming removes the inflectional suffixes in a word (e.g., working $\rightarrow$ work).
The contraction normalization expands one contracted word to two separate words (e.g., didn't $\rightarrow$ did n't), aligning with the treebank's tokenization.

Thirdly, we compute the span-overlap score as the frequency of the normalized $w^{i, j}$ within the normalized sentence set $s$ (Equation~\ref{eq:span_overlap}).
The span-overlap score is an exact matching score, requiring $w^{i, j}$ to be a substring of sentences $s_k\in s$.
We also present results in appendix~\ref{sec:appendix-inexact} for a span-overlap variant using fuzzy matching scores.

% \vspace{-0.2cm}
\begin{equation}
    \small
    {\rm score}(w, (i, j)):= \frac{\sum_{s_k \in s} \mathbbm{1} (w^{i, j} \text{ is a substring in } s_k)}{|s|}
    \label{eq:span_overlap}
\end{equation}

Lastly, we decode the constituent structure by finding a tree with maximum span-overlap scores.
We apply the Viterbi algorithm \cite{1090714} for this purpose.

\section{Experiment}
\label{sec:experiment}
\begin{table*}[t]
    \small
    \centering
    % \vspace{-0.8cm}
    \adjustbox{width=0.9\textwidth}{
        \begin{tabular}{|l|llllllll|l|}
            \hline
            SPMRL        & German         & French        & Hungarian      & Swedish        & Polish         & Basque         & Hebrew        & Korean         & Mean           \\\hline
            % Random Tree     & 13.9           & 16.2          & 14.1           & 16.4           & 21.4           & 19.5           & 19.7          & 22.2           & 17.9           \\
            % Left Branching  & 10.0           & 5.7           & 13.3           & 8.4            & 10.9           & 17.9           & 8.5           & 18.5           & 11.6           \\
            % Right Branching & 14.7           & 26.4          & 12.7           & 30.4           & 34.2           & 15.4           & 30.0          & 19.2           & 22.8           \\\hline
            % GPT 3.5*     & -     & -    & 28.7     & 35.2    & 40.5   & 21.1    & 2.1    & 4.0    & -     \\\hline
            GPT-3.5 & 29.4&25.0&28.7&35.2&40.5&21.1&2.1&4.0&23.2 \\\hline
            N-PCFG\dag   & 37.8           & 42.2          & 37.9           & 14.5           & 31.7           & 30.2           & 41.0          & 25.7           & 32.6           \\
            C-PCFG\dag   & 37.3           & 40.5          & 38.3           & 23.7           & 32.4           & 27.9           & 39.2          & 27.7           & 33.3           \\
            TN-PCFG\dag  & 47.1           & 39.1          & 43.1           & 40.0           & 48.6           & 36.0           & 39.2          & 35.4           & 40.8           \\\hline
            MLM-IP\ddag  & 40.8           & \textbf{48.7} & 39.1           & 46.3           & 53.3           & 44.0           & \textbf{50.4} & 43.7           & 45.7           \\
            Span Overlap & \textbf{49.5} & 48.5         & \textbf{48.1} & \textbf{51.2} & \textbf{61.6} & \textbf{49.7} & 43.8         & \textbf{48.1} & \textbf{50.1} \\\hline
        \end{tabular}}
    \caption{SF1 score on SPMRL's test set. The best results are highlighted in bold. \dag: Results from \cite{yang-etal-2021-pcfgs}. \ddag: Results from \cite{li-lu-2023-contextual}.}
    \label{tbl:spmrl-comparison}
    \vspace{-0.4cm}
\end{table*}

\subsection{Experimental Settings}
We conduct experiments in 10 languages, namely English (Penn Treebank (PTB)) \cite{Treebank-3}, Chinese (Penn Chinese Treebank 5.1 (CTB)) \cite{CTB5.1}\footnote{The PTB and CTB datasets are available under the \texttt{LDC User Agreement for Non-Members} license.}, German, French, Hungarian, Swedish, Polish, Basque, Hebrew, and Korean (SPMRL) \cite{seddah-etal-2013-overview}.
We apply the same training/development/testing split as \citet{li-lu-2023-contextual}. 
We use the development section for analysis and the testing section for comparison with existing unsupervised parsers.
As per previous research \cite{DBLP:conf/iclr/ShenLHC18, kim-etal-2019-compound}, we evaluate the parsing accuracy using the unlabeled sentence-F1 (SF1) score on sentences of length $\geq$ 2.
We drop punctuation (e.g., comma and period) and trivial spans (i.e., spans containing a single word and the sentence level span) when computing the SF1 score.

We generate PAS-equivalent sentences using an instruction-following LLM \texttt{GPT-3.5-turbo-0125}.
We apply English instructions for all languages but add a specific directive asking the LLM to generate native samples for non-English languages.
We manually examined 100 samples in English and Chinese and found that most samples had PAS identical to or similar to the target sentence.\footnote{More details in Appendix~\ref{sec:appendix-samplequality}}
We apply the snowball-stemmer\footnote{Available as a part of the NLTK package.} for English, French, German, Basque, Hungarian, and Swedish, and the stempel-stemmer\footnote{Available as a part of \href{https://www.egothor.org/product/egothor2/index.html}{the egothor project.}} for Polish.
We skip the stemming process for Chinese, Korean, and Hebrew.
Chinese words do not take inflections.
Hebrew and Korean do not have a readily available stemmer.
We normalize contractions only in English and French because of the common use of contractions in these languages.
While not implementing the full normalization may harm the parsing accuracy in many languages, the span-overlap parser achieves state-of-the-art level performance without the full normalization in those languages.

We analyze the span-overlap score for two types of constituents: participant-denoting and event-denoting constituents.
The span-overlap score measures the word sequence frequency within the PAS-equivalent sentence set.
The objective is to identify how the language model behaves when generating constituents of different characteristics.
We refer to a constituent as participant-denoting when it denotes a participant in the PAS.
Likewise, we refer to a constituent as event-denoting when it denotes an event in the PAS.
For example, Noun Phrases (NP) and Prepositional Phrases (PP) are common participant-denoting constituents, whereas Verb Phrases (VP) and Inflectional Phrases (IP) are common event-denoting constituents.
We compare the span-overlap score for phrase types with more than twenty occurrences.
We normalize the phrase type annotation for three languages (French, Polish, and Basque) whose annotation does not align with the PTB's annotation.
In French, we merge the \texttt{Srel}, \texttt{Sint}, and \texttt{Ssub} phrases into S, and the \texttt{VPpart} and \texttt{VPinf} into VP.
In Polish, we treat the \texttt{zdanie} phrase as S, the \texttt{fwe} as VP, and the \texttt{fno} as NP \cite{wolinski-etal-2018-new}.
In Basque, we treat the \texttt{SN} phrase as NP and the \texttt{SP} phrase as PP \cite{DBLP:journals/pdln/SanchezFAA08}.

\subsection{Comparison with the State-of-the-Art}

\begin{table}[t]
    \small
    \centering
    % \vspace{-0.2cm}
    \adjustbox{width=0.7\columnwidth}{
        \begin{tabular}{|l|ll|}
            \hline       & English              & Chinese       \\\hline
            GPT 3.5     & 36.2       & 5.1    \\\hline
            N-PCFG\dag   & 50.8      (50.8)     & 29.5  (34.7)  \\
            C-PCFG\dag   & 55.2  (52.6)         & 39.8   (37.3) \\
            TN-PCFG\dag  & \textbf{57.7} (51.2) & 39.2 (39.3)   \\\hline
            MLM-IP\ddag  & 49.0 (46.4)          & -   (37.8)    \\
            Span Overlap & 52.9                & \textbf{48.7} \\\hline
        \end{tabular}}
    \caption{SF1 score on the English (PTB) and the Chinese (CTB) test set. The best results are highlighted in bold. \dag: results from \cite{yang-etal-2021-pcfgs}; \ddag: results from \cite{li-lu-2023-contextual}; -: result not reported. Numbers in parentheses () are our reproduction with the released code and data. }
    % Reproduction results for the PCFG-based methods are the average of four independent runs.}
    \label{tbl:ptbnctb-comprison}
    \vspace{-0.4cm}
\end{table}

Table~\ref{tbl:spmrl-comparison} and Table~\ref{tbl:ptbnctb-comprison} showcase the effectiveness and applicability of the frequency information revealed in the PAS-equivalent sentence set.
The two tables compare the span-overlap parser with three PCFG-based parsers (N-PCFG, C-PCFG \cite{kim-etal-2019-compound}, and TN-PCFG \cite{yang-etal-2021-pcfgs}) and the latest language model-based parser (MLM-IP \cite{li-lu-2023-contextual}).
We also include a GPT-3.5 parser that predicts PTB style parse tree in its serialized form (e.g., ``(S (NP (DT The) (NN value)) (VP (VBD was) (VP (VBN disclosed))))'').
We drop trivial and duplicate spans when computing the unlabelled SF1 score for the GPT-3.5 parser.

The span-overlap parser represents a significant improvement over previous unsupervised parsers, outperforming the best baseline parser in eight out of ten languages.
The span-overlap parser outperforms the MLM-IP parser by 3.99 SF1, 10.9 SF1, and 4.32 mean SF1 scores in the English PTB, the Chinese CTB, and the multilingual SPMRL dataset, respectively.
It respectively outperforms the best grammar-based parser (TN-PCFG) by 9.5 SF1 and 9.31 mean SF1 scores in the Chinese CTB and the multilingual SPMRL dataset but underperforms the TN-PCFG parser by 4.7 SF1 score in the English PTB dataset.
The high parsing accuracy establishes the span-overlap parser as a simple but competitive multilingual unsupervised parser.

The strong parsing performance is not solely due to the application of the \texttt{GPT-3.5} model.
As shown in the table, the GPT-3.5 parser is the worst-performing baseline parser in the PTB, CTB, and SPMRL dataset, achieving 36.2 SF1 in the PTB, 5.1 SF1 in the CTB, and 23.2 mean SF1 score in the SPMRL dataset.
The low parsing accuracy indicates that the GPT-3.5 model can only recover a small portion of the constituent structure.
In contrast, the proposed span-overlap parser uncovers significantly more constituent structures, outperforming the GPT-3.5 parser by 16.7 SF1 in the PTB, by 43.6 SF1 in the CTB, and by 26.86 mean SF1 score in the SPMRL dataset.

\begin{figure*}[ht]
    \centering
    % \vspace{-1.2cm}
    \includegraphics[width=\textwidth]{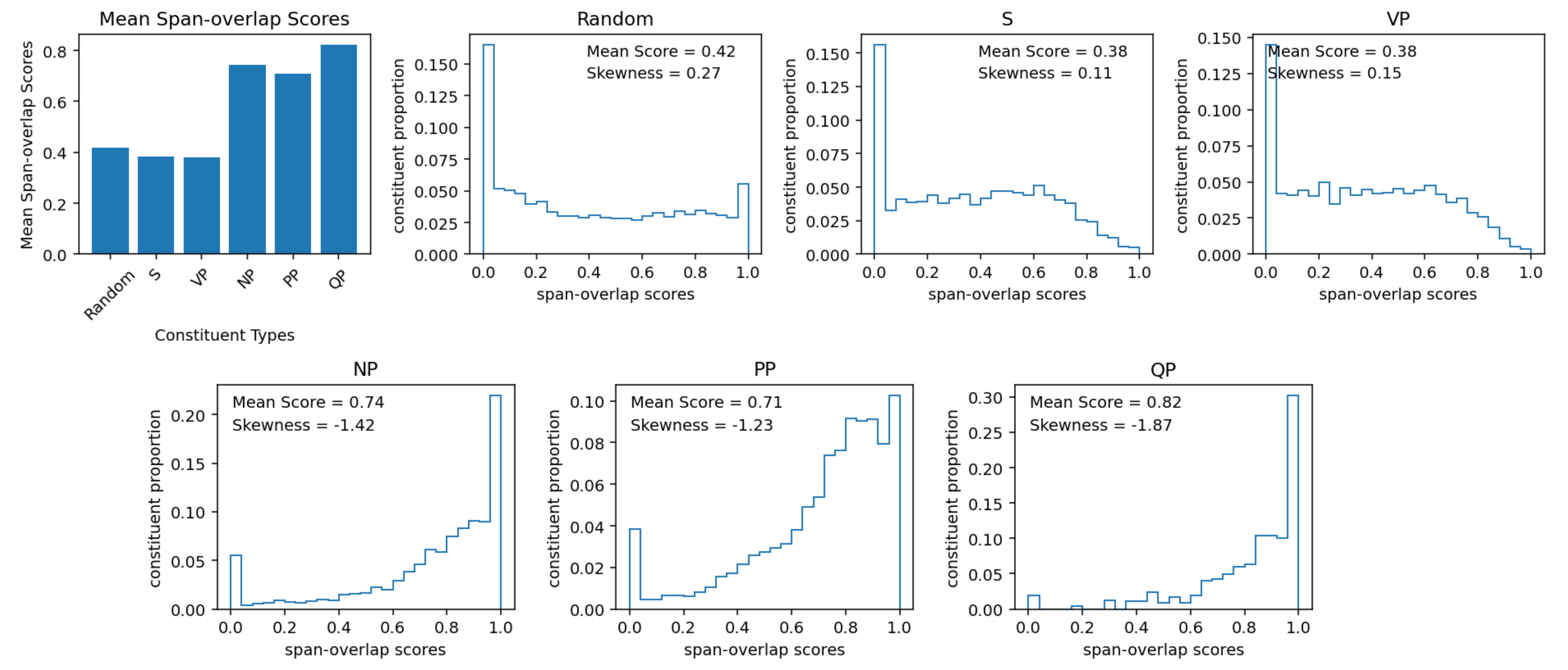}
    % \vspace{-0.4cm}
    \caption{The mean span-overlap score and their distributions for S, VP, NP, PP, QP, and Random in the PTB development set. The distribution is approximated with histograms.}
    \label{fig:mean_scores'ndistributions}
    % \vspace{-0.4cm}
\end{figure*}

\begin{figure}[h]
    \centering
    \begin{subfigure}[b]{0.22\textwidth}
        \centering
        \includegraphics[width=\textwidth]{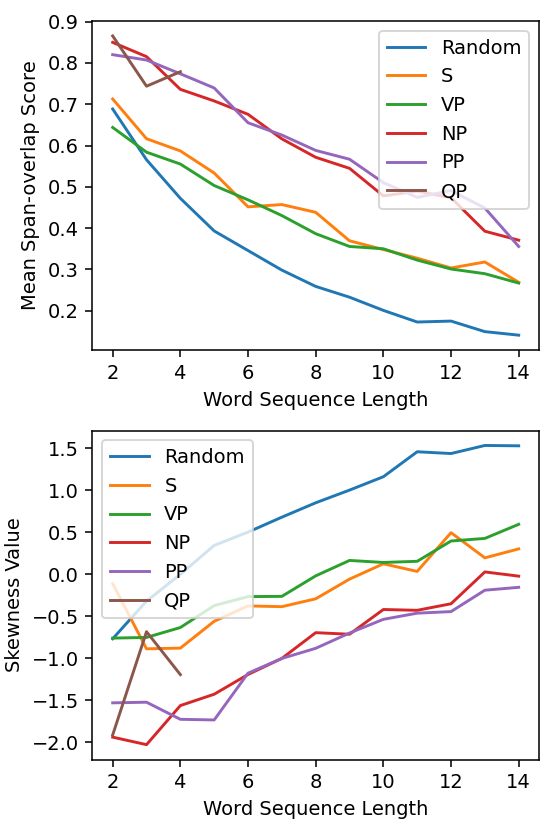}
        % \vspace{-0.6cm}
        \caption{English}
    \end{subfigure}
    % \hfill
    \begin{subfigure}[b]{0.22\textwidth}
        \centering
        \includegraphics[width=\textwidth]{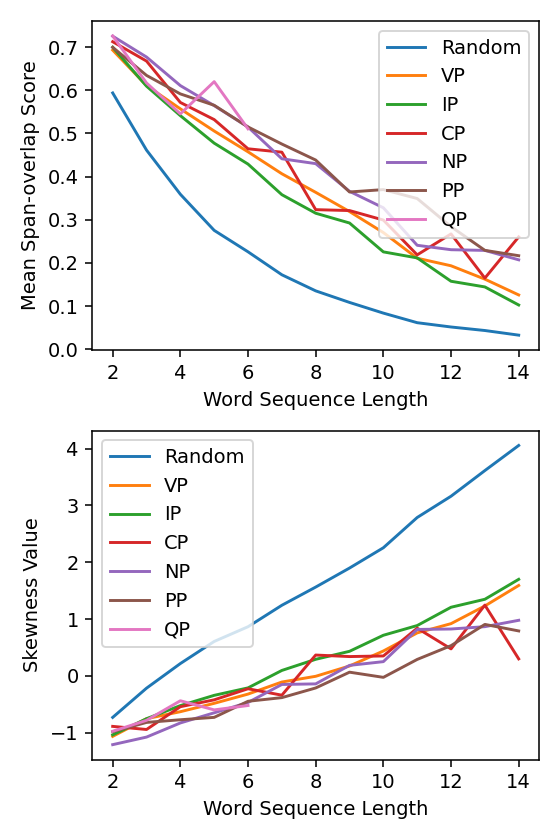}
        % \vspace{-0.6cm}
        \caption{Chinese}
    \end{subfigure}
    \hfill
    % \vspace{-0.3cm}
    \caption{Mean and skewness values of span-overlap scores. }
    % \vspace{-0.5cm}
    \label{fig:mean'nskewness_by_spanlen}
\end{figure}

\subsection{Participant-denoting Constituents Tend to Have Higher Span-overlap Scores than Equal-length Event-denoting Constituents}
In this section, we demonstrate a statistical difference between equal-length participant-denoting and event-denoting constituents through the span-overlap score.
We focus on four participants-denoting constituents (i.e., Noun Phrases (NP), Prepositional Phrases (PP), Adpositional Phrases (AP), and Quantifier Phrases (QP)) and four event-denoting constituents (i.e., Declarative Clauses (S), Verb Phrases (VP), Inflectional Phrases (IP), and Complementizer Phrases (CP)).
We also include the span-overlap score for random word sequences with the same sequence length statistics as gold constituents (i.e., For each random sequence, there is a gold constituent with the same sequence length).

We begin by observing the difference between the two constituent types regardless of the constituent's sequence length.
Figure~\ref{fig:mean_scores'ndistributions} shows the mean span-overlap score along with the score distribution for NP, PP, QP, S, and VP in the English PTB dataset.
The AP, IP, and CP are not annotated in the dataset and, therefore, cannot be shown in the figure.
The three participant-denoting constituents (i.e., NP, PP, and QP) have higher mean scores than the two event-denoting constituents (i.e., S and VP).
The distribution figure shows that most participant-denoting constituents have high scores, whereas most event-denoting constituents do not.
We can characterize this feature using the skewness value of the score distribution \cite{Russo2020}.
A lower skewness value indicates more constituents with high scores than those with low scores.
As shown in the figure, the participant-denoting constituents have a negative skewness value, whereas the event-denoting constituents have a positive skewness value.
This skewness value indicates that participant-denoting constituents tend to appear more frequently in the PAS-equivalent sentence set than event-denoting constituents.
The comparison establishes the use of the mean score and skewness value as a metric to reflect the statistical difference between the two constituent types.

Figure~\ref{fig:mean'nskewness_by_spanlen} demonstrates the statistical difference between equal-length participant-denoting and event-denoting constituents in English and Chinese.
Comparing constituents of different lengths, we see that short constituents tend to have higher scores than long constituents.
The correlation between span-overlap scores and sequence lengths prompts whether the statistical difference is due to the length difference.
This is unlikely the case.
As shown in the figure, participant-denoting constituents have higher mean span-overlap scores and lower skewness values than equal-length event-denoting constituents \emph{over all sequence lengths}.
In other words, participant-denoting constituents tend to appear more frequently in the PAS-equivalent sentences than equal-length event-denoting constituents.
We include the mean score and skewness value for the other eight languages in Figure~\ref{fig:mean'nskewness_by_spanlen-appendix}, which shows the same trend as in Chinese and English.
These results highlight a multilingual phenomenon: participant-denoting constituents tend to have higher span-overlap scores than equal-length event-denoting constituents.
The phenomenon indicates a statistical difference between the two constituent types, shedding light on future unsupervised parsing research with phrase labels.

\subsection{Span-overlap Score Distinguishes between Constituents and Non-constituents}

\begin{figure*}[t]
    % \vspace{-1.2cm}
    \centering
    \begin{subfigure}[b]{0.19\textwidth}
        \centering
        \includegraphics[width=\textwidth]{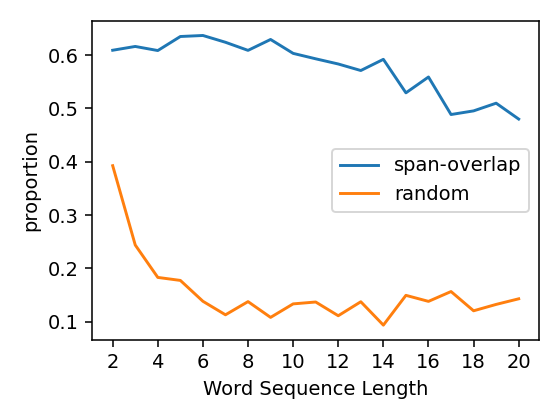}
        % \vspace{-0.8cm}
        \caption{English}
    \end{subfigure}
    % \hfill
    \begin{subfigure}[b]{0.19\textwidth}
        \centering
        \includegraphics[width=\textwidth]{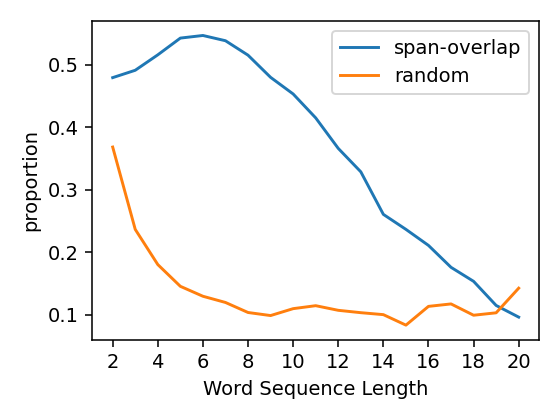}
        % \vspace{-0.8cm}
        \caption{Chinese}
    \end{subfigure}
    \hfill
    \begin{subfigure}[b]{0.19\textwidth}
        \centering
        \includegraphics[width=\textwidth]{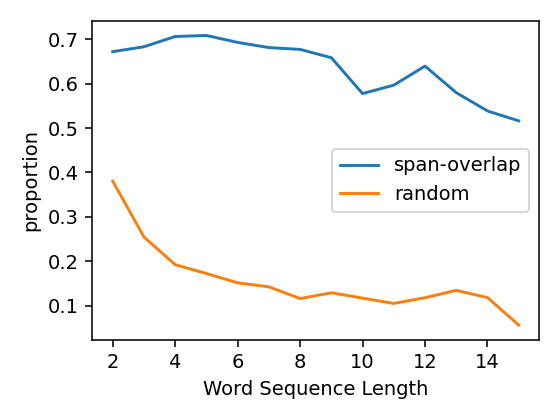}
        % \vspace{-0.8cm}
        \caption{German}
    \end{subfigure}
    % \hfill
    \begin{subfigure}[b]{0.19\textwidth}
        \centering
        \includegraphics[width=\textwidth]{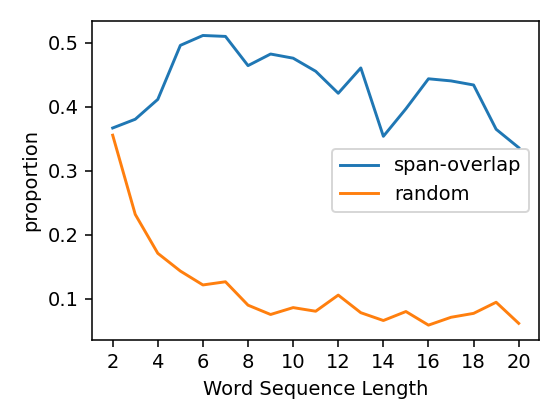}
        % \vspace{-0.8cm}
        \caption{French}
    \end{subfigure}
    \hfill
    \begin{subfigure}[b]{0.19\textwidth}
        \centering
        \includegraphics[width=\textwidth]{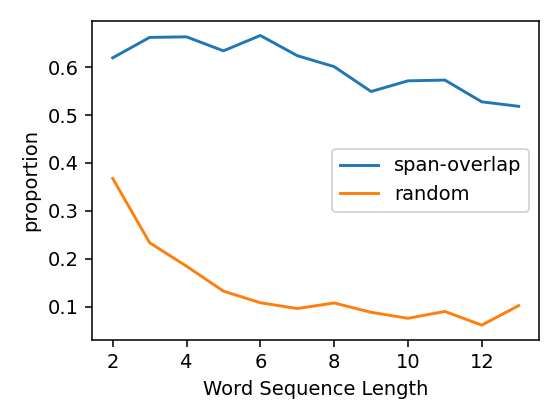}
        % \vspace{-0.8cm}
        \caption{Hungarian}
    \end{subfigure}
    % \hfill
    \begin{subfigure}[b]{0.19\textwidth}
        \centering
        \includegraphics[width=\textwidth]{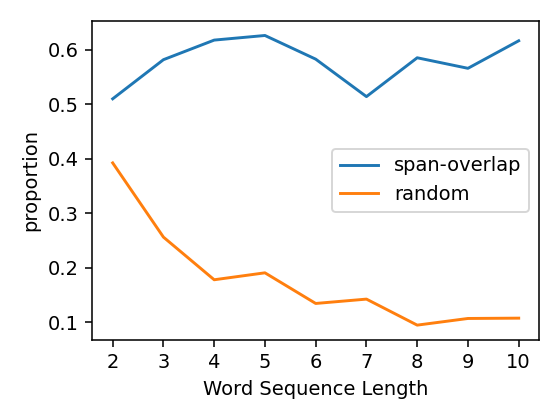}
        % \vspace{-0.8cm}
        \caption{Swedish}
    \end{subfigure}
    \hfill
    \begin{subfigure}[b]{0.19\textwidth}
        \centering
        \includegraphics[width=\textwidth]{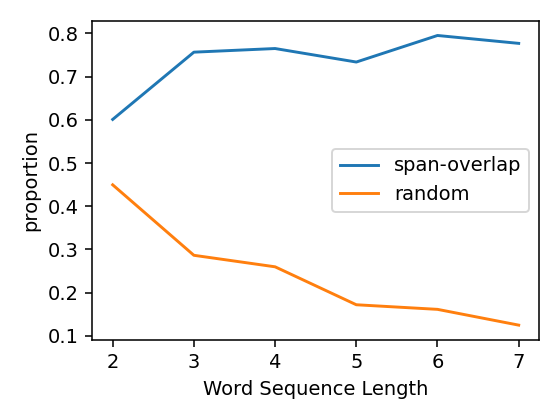}
        % \vspace{-0.8cm}
        \caption{Polish}
    \end{subfigure}
    % \hfill
    \begin{subfigure}[b]{0.19\textwidth}
        \centering
        \includegraphics[width=\textwidth]{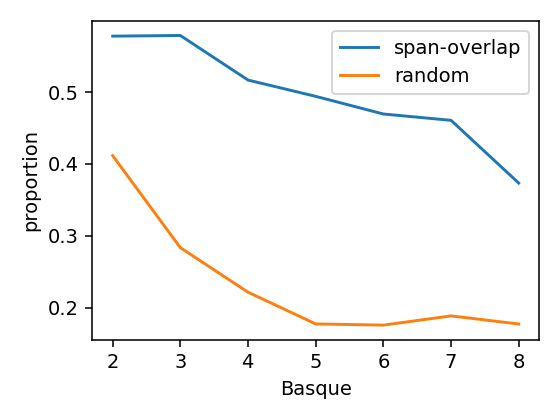}
        % \vspace{-0.8cm}
        \caption{Basque}
    \end{subfigure}
    \hfill
    \begin{subfigure}[b]{0.19\textwidth}
        \centering
        \includegraphics[width=\textwidth]{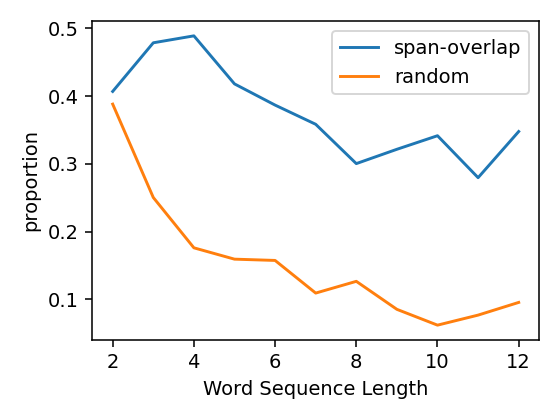}
        % \vspace{-0.8cm}
        \caption{Hebrew}
    \end{subfigure}
    % \hfill
    \begin{subfigure}[b]{0.19\textwidth}
        \centering
        \includegraphics[width=\textwidth]{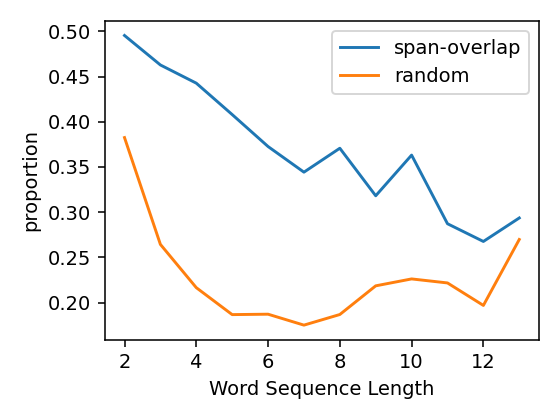}
        % \vspace{-0.8cm}
        \caption{Korean}
    \end{subfigure}
    \hfill
    % \vspace{-0.2cm}
    \caption{Proportion of gold constituents to which the scores assign strictly higher values than non-constituents}
    % \vspace{-0.3cm}
    \label{fig:proportions}
\end{figure*}

\begin{table*}[ht]
    \centering
    \small
    \adjustbox{width=\textwidth}{
        \begin{tabular}{|l|l|l|l|l|l|l|l|l|l|l|}
            \hline
            Ablation              & English    & Chinese    & German      & French     & Hungarian  & Swedish    & Polish      & Basque     & Hebrew     & Korean     \\ \hline
            Random                & 13.33      & 21.13      & 24.50        & 12.45      & 9.91       & 14.87      & 26.83       & 16.90       & 18.48      & 22.59      \\ \hline
            Right-branching       & 37.61      & 29.79      & 28.14       & 27.62      & 11.24      & 35.55      & 37.33       & 13.63      & 31.37      & 16.59      \\ \hline
            GPT-3.5              & 36.20 & 5.50 & 29.43 & 25.04 & 19.07 & 39.88 & 39.21 & 17.52 & 2.20 & 4.90 \\ \hline\hline
            Movement              & 53.08      & 44.39      & 48.55       & 48.55      & 34.68      & 48.63      & 59.77       & 42.50       & 42.59      & 40.50       \\ \hline
            Question Conversion   & 50.95      & 40.90       & 42.31       & 42.78      & 34.01      & 45.25      & 53.29       & 34.96      & 38.57      & 35.09      \\ \hline
            Tense Conversion      & 42.93      & 36.81      & 33.05       & 30.53      & 19.65      & 37.20       & 44.40        & 18.77      & 35.83      & 24.22      \\ \hline
            Clefting              & 50.21      & 0          & 0           & 0          & 0          & 0          & 0           & 0          & 0          & 0          \\ \hline
            Passivization         & 52.15      & 37.09      & 39.11       & 42.65      & 26.57      & 41.20       & 51.47       & 25.11      & 39.92      & 29.06      \\ \hline
            Language-Specific     & 0          & 35.95      & 43.22       & 0          & 29.73      & 0          & 50.59       & 34.29      & 37.13      & 36.23      \\ \hline\hline
            Best Instruction Type & 53.08      & 44.39      & 48.55       & 48.55      & 34.68      & 48.63      & 59.77       & 42.50       & 42.59      & 40.50       \\ \hline
            Full Instructions     & 53.98      & 48.53      & 50.47       & 49.26      & 43.48      & 49.44      & 60.15       & 45.43      & 43.28      & 50.15      \\ \hline
            % Delta                 & 0.9     & 4.14    & 1.92   & 0.71   & 8.8       & 0.81    & 0.38   & 2.93   & 0.69   & 9.65   \\\hline
        \end{tabular}}

    % \end{tabular}}
    \caption{SF1 scores for span-overlap parser with only one instruction type and with all 6 instruction types. Results are obtained using the development set. }
    \label{tbl:ablation}
    % \vspace{-0.5cm}
\end{table*}

This experiment provides a glimpse of the discriminating power of the span-overlap score.
We compare the span-overlap score with random scores in terms of whether the score can identify the constituent in a set of mutually incompatible word sequences.
The random score is uniformly sampled from [0, 1], the same range as the span-overlap score.
We measure the proportion of gold constituents to which the score assigns a strictly higher value than all incompatible non-constituents.
We refer to a non-constituent as incompatible with a gold constituent when the two sequences partially overlap (i.e., they cannot co-exist in the same tree).

Figure~\ref{fig:proportions} plots the proportion by word sequence lengths.
The span-overlap score distinguishes between constituents and non-constituents, identifying more constituents than the random score for all sequence lengths and in all languages.
In the European languages, the span-overlap score reliably identifies around 50\% of constituents, regardless of the constituents' length.
In contrast, the random score identifies around 40\% of short constituents but only around 10\% of long constituents.
The comparison highlights the effectiveness of the span-overlap score in identifying constituents, especially for long constituents.
However, Chinese and Koreans are exceptions.
While the span-overlap score still identifies more constituents than the random score, the proportion drops for long constituents. 
This is because no long word sequences appear in both the target sentence and the generated PAS-equivalent sentences.
The problem might be related to the sample quality problem discussed in Section~\ref{sec:limitations}.

\subsection{Ablation of PAS-Preserving Instructions}

Table~\ref{tbl:ablation} showcases the benefit of aggregating multiple PAS-preserving instructions.
The table compares the span-overlap parser with only one instruction type and with all six instruction types.
Even with a single instruction type, the span-overlap parser outperforms the random, right-branching, and GPT-3.5 parsers.
This result suggests that the span-overlap method uncovers non-trivial constituent structures.
The best parser with a single instruction type is always the one with the movement instruction.
The phenomenon suggests a fundamental role of movable constituents in identifying constituent structures.
Combining all instruction types sees significant improvements in parsing accuracy.
The average improvement is 3.09 SF1 score between the parser with all instruction types and the best parser with a single instruction type.
This result indicates the benefit and necessity of aggregating multiple instructions in generating the PAS-equivalent sentence set.

\section{Conclusion}
In this paper, we approached the unsupervised constituency parsing problem by exploiting word sequence frequency information revealed in a set of PAS-equivalent sentences.
Firstly, we empirically verified a hypothesis: \textbf{constituents correspond to frequent word sequences in the PAS-equivalent sentence set}.
We proposed a frequency-based method \emph{span-overlap} that computes the span-overlap score as the word sequence's frequency in the PAS-equivalent sentence set.
The span-overlap parser outperformed state-of-the-art parsers in eight out of ten languages evaluated and significantly outperformed the GPT-3.5 parser in all languages.
The span-overlap score can effectively separate constituents from non-constituents, especially for long constituents.
These results highlight the effectiveness and applicability of the frequency information to unsupervised parsing.
Additionally, our study uncovers a \emph{multilingual} phenomenon: participant-denoting constituents tend to have higher span-overlap scores than equal-length event-denoting constituents, meaning that the former tend to appear more frequently in the PAS-equivalent sentence set than the latter. 
The phenomenon indicates a statistical difference between the two constituent types, laying the foundation for future labeled unsupervised parsing research.
Finally, our ablation analysis shows a significant improvement by aggregating multiple PAS-preserving transformations, indicating the necessity of the aggregation in comprehensively exposing the constituent structure.

\section{Limitations}
\label{sec:limitations}
The span-overlap method employs a diverse but non-exhaustive set of instructions to generate PAS-equivalent sentences.
We did not cover PAS-preserving transformations that are unique to some languages.
For example, Korean and Chinese have a null subject transformation that can remove the subject from sentences.
The span-overlap method may miss out on word sequence patterns that those transformations cover.
However, this paper aims to demonstrate the utility of word sequence frequency information and verify that constituents correspond to frequent word sequences.
We will leave the incorporation of more complex language-specific transformations to future work.

Additionally, the span-overlap method faces challenges with sample quality, particularly in non-English languages.
Previous research \cite{Rosl2023EvaluationOT, Seghier2023ChatGPTNA} has pointed out that the \texttt{GPT-3.5} model performs significantly better in English than in other languages.
Similarly, we observed that English samples are of better quality than Chinese samples.
We refer the readers to Appendix~\ref{sec:appendix-samplequality} for more details.
We also observed that using \texttt{GPT-4} instead of \texttt{GPT-3.5} can significantly improve the parsing accuracy in a preliminary experiment.
These results suggest that the sample quality currently limits the span-overlap parser.
That being said, the span-overlap parser's performance indicates the utility of the word sequence frequency estimated with even low-quality samples.
We expect the sample quality issue to be mitigated in the future with the rapid development of LLM technologies.

% \citet{eriguchi-etal-2019-incorporating}

% Bibliography entries for the entire Anthology, followed by custom entries
\bibliography{anthology, anthology_p2, custom}
% Custom bibliography entries only
% \bibliography{custom}

\appendix

\begin{table*}[t]
    \centering
    \small
    \begin{tabular}{|l|l|}
        \hline
                                             & Prompts                                                                                                           \\ \hline
        \multirow{4}{*}{Movement}            & \shortstack[l]{Generate diverse sentences by performing the topicalization transformation to the below sentence.} \\ \cline{2-2}
                                             & \shortstack[l]{Generate diverse sentences by performing the preposing transformation to the below sentence.}      \\ \cline{2-2}
                                             & \shortstack[l]{Generate diverse sentences by performing the postposing transformation to the below sentence.}     \\ \cline{2-2}
                                             & \shortstack[l]{Generate diverse sentences by performing the extraposition transformation to the below sentence.}  \\ \cline{1-2}
        \multirow{2}{*}{Question Conversion} & \shortstack[l]{Generate diverse sentences by converting the below sentence to a yes-no question.}                 \\ \cline{2-2}
                                             & \shortstack[l]{Generate diverse sentences by converting the below sentence to a wh-question.}                     \\ \hline
        Tense Changing                       & \shortstack[l]{Generate diverse sentences by converting the below sentence to other tenses.}                      \\ \hline
        Passivization                        & \shortstack[l]{Generate diverse sentences by passivizing the below sentence.}                                     \\ \hline
        Clefting                             & \shortstack[l]{Generate diverse sentences by converting the below sentence to a cleft sentence. }                 \\ \hline
        \multirow{3}{*}{Question Conversion} & \shortstack[l]{Generate diverse sentences by performing the scrambling transformation to the below sentence. }    \\ \cline{2-2}
                                             & \shortstack[l]{Generate diverse sentences by performing heavy NP shift. }                              \\ \cline{2-2}
                                             & \shortstack[l]{Generate diverse sentences by converting the below sentence into ones with ba-construction.}       \\ \hline
    \end{tabular}
    \caption{Instructions used in generating PAS-equivalent sentences. }
    \label{tbl:PAS-equivalent-instructions}
\end{table*}

\begin{figure*}[!ht]
    \centering
    % \begin{subfigure}[b]{0.24\textwidth}
    %     \centering
    %     \includegraphics[width=\textwidth]{figs/english_dev.png}
    %     % \vspace{-0.6cm}
    %     \caption{English}
    % \end{subfigure}
    % % \hfill
    % \begin{subfigure}[b]{0.24\textwidth}
    %     \centering
    %     \includegraphics[width=\textwidth]{figs/chinese_dev.png}
    %     % \vspace{-0.6cm}
    %     \caption{Chinese}
    % \end{subfigure}
    % \hfill
    \begin{subfigure}[b]{0.24\textwidth}
        \centering
        \includegraphics[width=\textwidth]{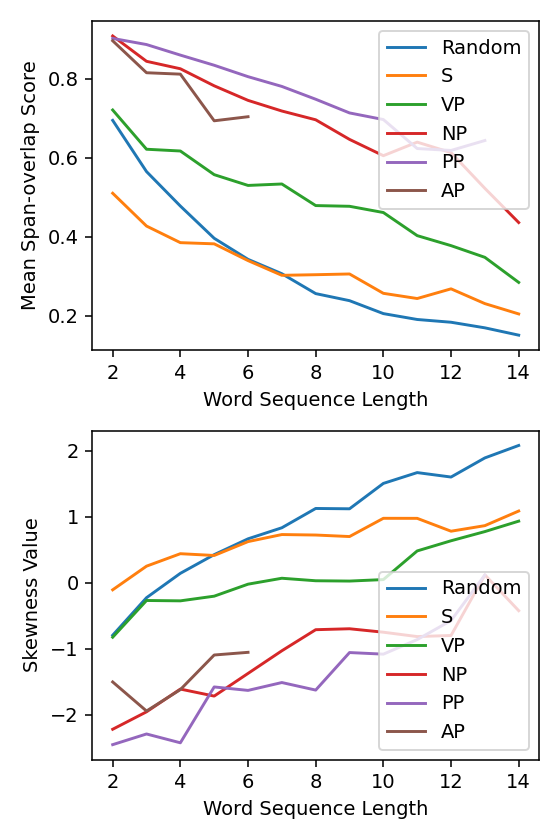}
        % \vspace{-0.6cm}
        \caption{German}
    \end{subfigure}
    % \hfill
    \begin{subfigure}[b]{0.24\textwidth}
        \centering
        \includegraphics[width=\textwidth]{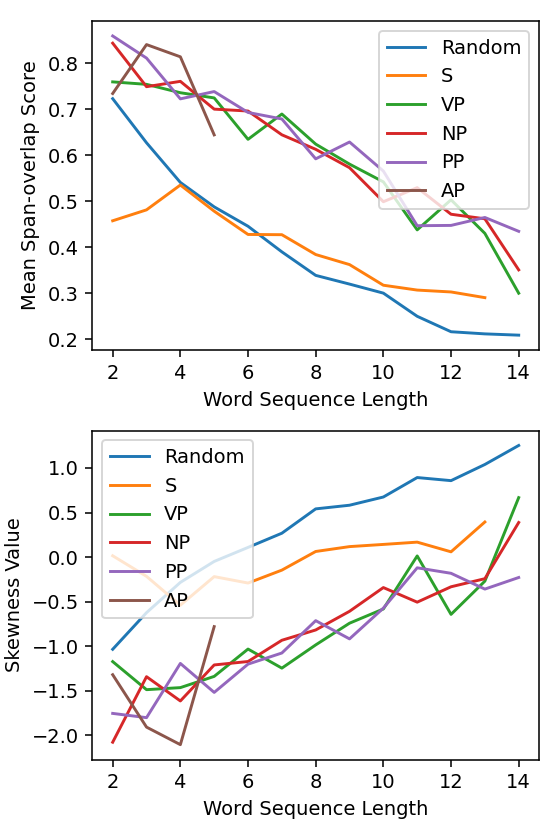}
        % \vspace{-0.6cm}
        \caption{French}
    \end{subfigure}
    \hfill
    \begin{subfigure}[b]{0.24\textwidth}
        \centering
        \includegraphics[width=\textwidth]{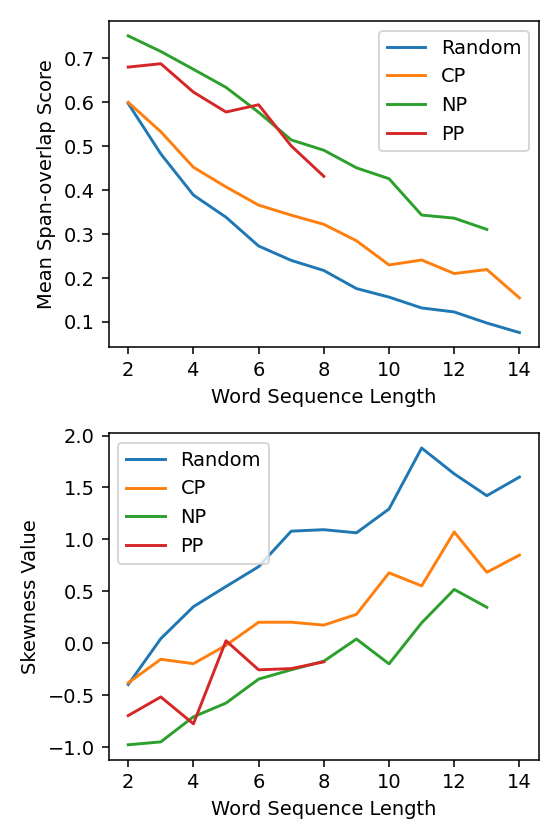}
        % \vspace{-0.6cm}
        \caption{Hungarian}
    \end{subfigure}
    % \hfill
    \begin{subfigure}[b]{0.24\textwidth}
        \centering
        \includegraphics[width=\textwidth]{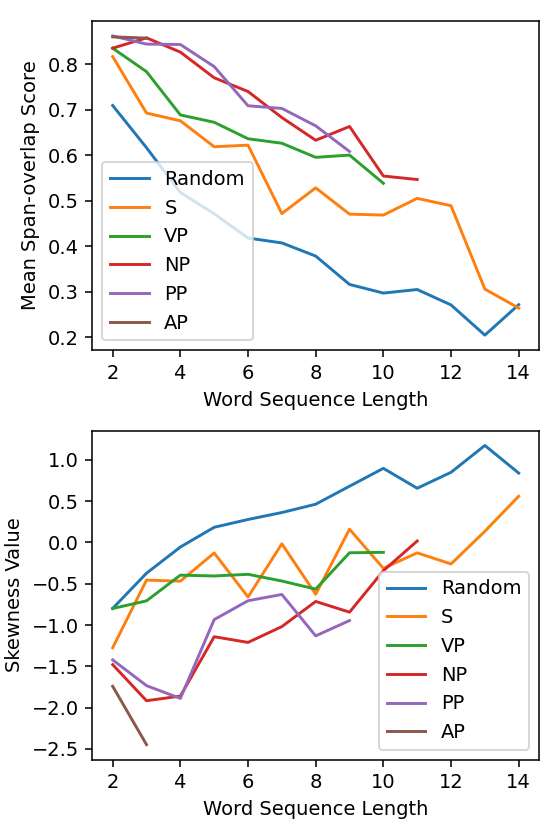}
        % \vspace{-0.6cm}
        \caption{Swedish}
    \end{subfigure}
    \hfill
    \begin{subfigure}[b]{0.24\textwidth}
        \centering
        \includegraphics[width=\textwidth]{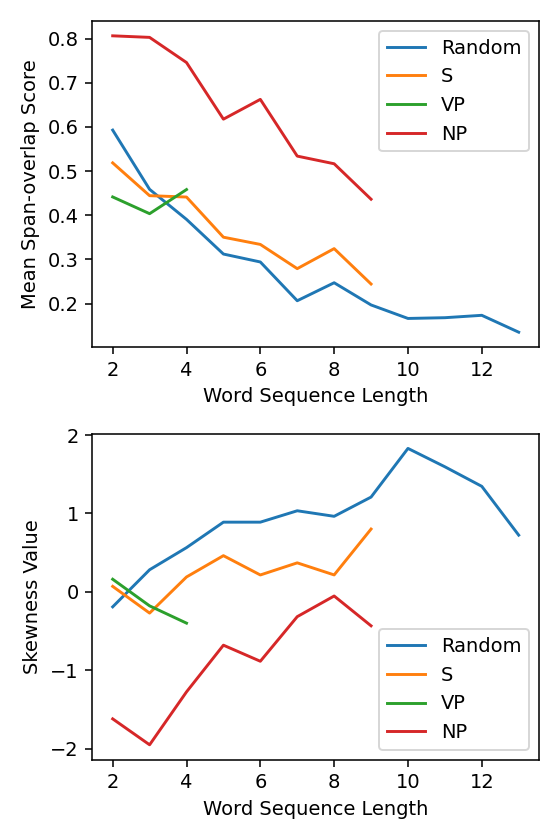}
        % \vspace{-0.6cm}
        \caption{Polish}
    \end{subfigure}
    % \hfill
    \begin{subfigure}[b]{0.24\textwidth}
        \centering
        \includegraphics[width=\textwidth]{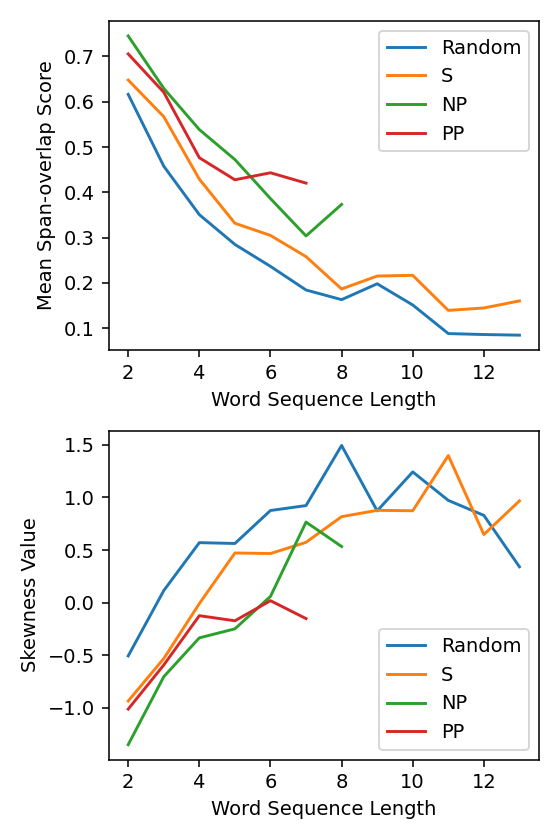}
        % \vspace{-0.6cm}
        \caption{Basque}
    \end{subfigure}
    \hfill
    \begin{subfigure}[b]{0.24\textwidth}
        \centering
        \includegraphics[width=\textwidth]{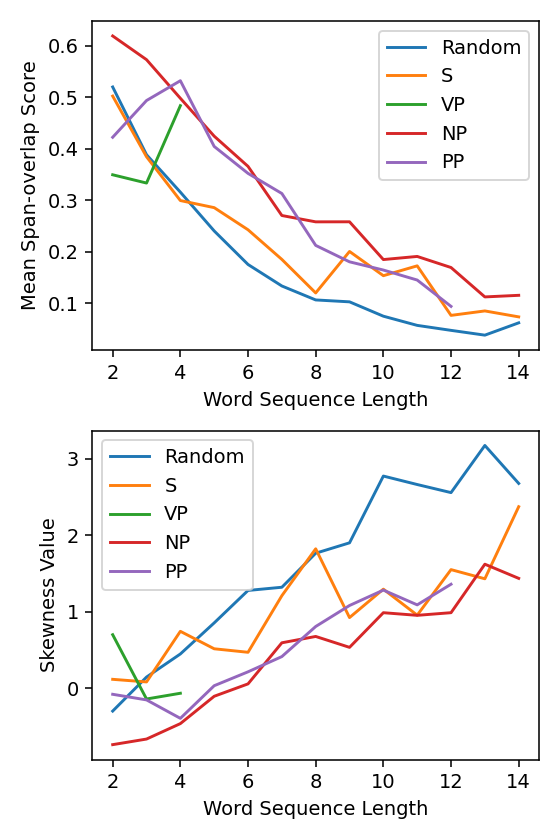}
        % \vspace{-0.6cm}
        \caption{Hebrew}
    \end{subfigure}
    % \hfill
    \begin{subfigure}[b]{0.24\textwidth}
        \centering
        \includegraphics[width=\textwidth]{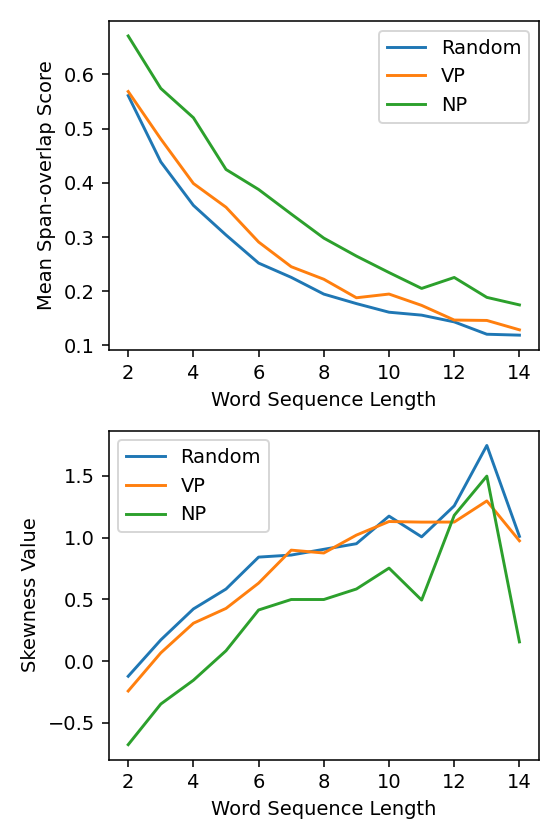}
        % \vspace{-0.6cm}
        \caption{Korean}
    \end{subfigure}
    \hfill
    % \vspace{-0.3cm}
    \caption{Mean and skewness values of span-overlap scores. The values are measured over development sets and are grouped by their sequence lengths. }
    % \vspace{-0.5cm}
    \label{fig:mean'nskewness_by_spanlen-appendix}
\end{figure*}

% \begin{figure}[h]
%     \centering
%     \begin{subfigure}[b]{0.22\textwidth}
%         \centering
%         \includegraphics[width=\textwidth]{figs/english_dev.png}
%         % \vspace{-0.6cm}
%         \caption{English}
%     \end{subfigure}
%     % \hfill
%     \begin{subfigure}[b]{0.22\textwidth}
%         \centering
%         \includegraphics[width=\textwidth]{figs/chinese_dev.png}
%         % \vspace{-0.6cm}
%         \caption{Chinese}
%     \end{subfigure}
%     \hfill
%     % \vspace{-0.3cm}
%     \caption{Mean and skewness values of span-overlap scores. The values are measured over development sets and are grouped by their sequence lengths. }
%     % \vspace{-0.5cm}
%     \label{fig:mean'nskewness_by_spanlen}
% \end{figure}

\section{Appendix}

\subsection{Span-overlap Scoring with Fuzzy Matching}
\label{sec:appendix-inexact}
The application of exact matching (i.e., the span only matches identical substrings in PAS-equivalent sentences) rather than fuzzy matching (i.e., the span matches similar substrings) requires further discussions.
In a preliminary experiment, we found that the fuzzy matching variants of the span-overlap parser underperform the span-overlap parser utilizing exact matching.
We experimented with two fuzzy matching variants, one based on the BLEU score \cite{papineni-etal-2002-bleu} and the other based on the METEOR score \cite{banerjee-lavie-2005-meteor}.
The two fuzzy matching scores allow the matching between two textually different word sequences that share the same meaning.
For example, ``give John a book'' and ``give a book to John''.

\begin{equation}
    \small
    \rm{score}(w, (i, j)):= \frac{\sum_{s_k \in s}\max_p f(w^{i,j}, s_k^{p,p+j-i})}{|s|}
    \label{eq:span_overlap-soft}
\end{equation}

\begin{table}[!ht]
    \centering
    \small
    \begin{tabular}{|l|l|l|l|}
        \hline
        ~               & METEOR & BLEU & EM   \\ \hline
        PTB-DEV-10      & 59.7   & 62.9 & 66.6 \\ \hline
        SPMRL-DE-DEV-10 & 63.7   & 64.4 & 67.9 \\ \hline
    \end{tabular}
    \caption{Sentence-F1 score on the 10-word subset of the PTB and SPMRL German development set.}
    \label{tbl:soft-scoring}
\end{table}

To introduce the fuzzy matching variants, we generalize Equation~\ref{eq:span_overlap} as Equation~\ref{eq:span_overlap-soft}.
When the scoring function $f$ is an indicator function, Equation~\ref{eq:span_overlap-soft} is the span-overlap score we introduced in the main text.
In the BLEU and METEOR score-based variants, we employ the BLEU and  METEOR scores for $f$, respectively.
Table~\ref{tbl:soft-scoring} compares the parsing accuracy of the two fuzzy matching variants against the exact matching variant on a 10-word subset of the development set of the PTB and the SPMRL German dataset.
The 10-word subset contains sentences of length $\leq10$, as per previous research \cite{klein-manning-2002-generative}.
As shown in the table, the fuzzy matching variants fall behind the exact matching variant in parsing accuracy by a large margin.
Upon further investigation, we found that fuzzy matching scores not only match ``give John a book'' with grammatical strings (e.g., ``give a book to John''), but also match with ungrammatical strings (e.g., ``give John a book in'').
The matching with ungrammatical strings might introduce unnecessary noise to the span-overlap model, which is detrimental to the parsing performance.

\subsection{PAS-preserving Instructions}
\label{sec:appendix-instru}
Table~\ref{tbl:PAS-equivalent-instructions} shows the specific instruction we used in our experiment.
To generate samples for non-English languages, we append an additional directive, ``The generated sentences must be in <language>.'', to the above instructions.
For example, we add ``The generated sentences must be in German.'' to generate German sentences.

\subsection{Examining Sample Qualities}
\label{sec:appendix-samplequality}
\begin{table*}[t]
    \centering
    \small
    \adjustbox{width=\textwidth}{
        \begin{tabular}{|l|l|l|}
            \hline
            Original Sentence                                                   & Generated Sentence                                                              & Rating \\ \hline
            Idle Thought                                                        & \shortstack[l]{The child, lost in the park, found solace in a world filled with          \\ color and wonder.} & 0      \\ \hline
            \shortstack[l]{BANKERS ACCEPTANCES : 8.40 \% 30 days ; 8.35 \% 60 days ;                                                                                       \\ 8.27 \% 90 days ;                                                                        8.20 \% 120 days ; 8.15 \% 150 days ; 8.02 \% 180 days .} & \shortstack[l]{Among the options are bankers acceptances at 8.02\% for 180 days,\\ catering to different investment durations.} & 6      \\ \hline
            They did n't panic during the first round of selling this morning . & They had not panicked during the first round of selling this morning.           & 10     \\ \hline
        \end{tabular}}
    \caption{Examples of PAS-equivalence rating}
    \label{tbl:rating-example}
\end{table*}

\begin{figure}
    \centering
    \includegraphics[width=0.8\columnwidth]{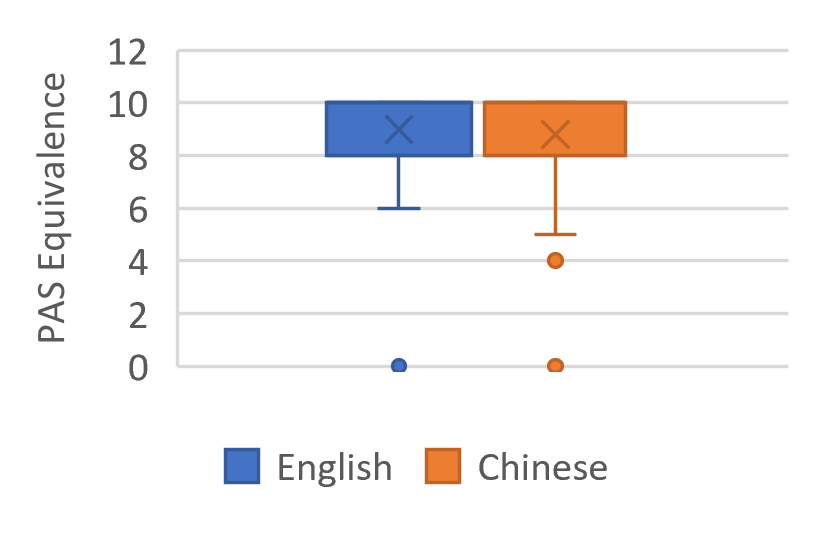}
    \caption{PAS-equivalence scores in English and Chinese}
    \label{fig:sample-quality}
\end{figure}

We manually examined 100 random pairs of (target sentence, generated PAS-equivalent sentence) in English and Chinese.
We rate the PAS-equivalence in the range of 0-10 (i.e., whether two sentences express the same PAS information).
We rate 10 if the two sentences express the same PAS information, 5 if half of the information from the target sentence is missing in the generated sentence, and 0 if the two sentences are completely irrelevant.
We present some rating examples in Table~\ref{tbl:rating-example}.

We can make two observations from Figure~\ref{fig:sample-quality}:
(1) Most generated sentences have a PAS that is reasonably similar to the PAS of the original sentence;
(2) English samples tend to have higher PAS-equivalence scores than Chinese samples.
These observations, on the one hand, confirm the plausibility of generating PAS-equivalent sentences with the \texttt{GPT-3.5} model.
On the other hand, the observations indicate a limiting factor for the span-overlap method: the sample quality.

\end{document}